\documentclass[lettersize,journal]{IEEEtran}
\usepackage{amsmath,amsfonts}
\usepackage{algorithmic}
\usepackage{algorithm}
\usepackage{array}
\usepackage[caption=false,font=normalsize,labelfont=sf,textfont=sf]{subfig}
\usepackage{textcomp}
\usepackage{stfloats}
\usepackage{url}
\usepackage{verbatim}
\usepackage{graphicx}
\usepackage{cite}
\usepackage{pifont}

\newcommand{\cmark}{\ding{51}}  
\newcommand{\xmark}{\ding{55}}  
\usepackage{tikz}
\usetikzlibrary{tikzmark}
\usetikzlibrary{positioning}

\begin{document}

\title{\LARGE \bf AI-IoT-Robotics Integration: Survey of Frameworks, Emerging Trends, and the Path Toward Connected Robotics}

\author{
Ranulfo Bezerra, Member, IEEE, Satoshi Tadokoro, Fellow, IEEE, Kazunori Ohno, Member, IEEE.
\thanks{This research was performed by the commissioned research fund provided by F-REI (JPFR23010101). (Corresponding author: Ranulfo Bezerra)}
\thanks{ 
R. Bezerra, S. Tadokoro, K. Ohno are with Tohoku University, Japan. E-mail: (ranulfo@tohoku.ac.jp; tadokoro@tr.is.tohoku.ac.jp; kazunori@tr.is.tohoku.ac.jp).
}
}


%


\maketitle

\begin{abstract}
The convergence of Artificial Intelligence, the Internet of Things, and Robotics is no longer a futuristic vision; it is rapidly becoming the foundation of real-time, intelligent, and context-aware systems. AI enables perception and reasoning, IoT provides scalable sensing and communication, and robotics delivers embodied actuation. Despite significant progress in pairwise combinations such as AIoT and the Internet of Robotic Things (IoRT), there remains a lack of unified design frameworks that fully integrate all three. This survey synthesizes the state-of-the-art across these domains, emphasizing the emerging role of Small Language Models (SLMs) at the edge and Large Language Models (LLMs) in the cloud for distributed cognition and autonomous decision-making. We propose a modular system architecture that aligns with these trends, analyze persistent gaps in interoperability and feedback control, and classify existing work by integration depth. Our review highlights how hybrid SLM–LLM systems, when coupled with IoT infrastructure and robotic agents, can address challenges in real-time adaptation, scalability, and reliability. This work offers a conceptual and technical roadmap for designing next-generation AI–IoT–Robotic ecosystems that are modular, interpretable, and capable of learning within dynamic environments, paving the way for the emerging paradigm of Connected Robotics and Physical AI.

\end{abstract}

%
\IEEEpeerreviewmaketitle

\section{Introduction}

The convergence of Artificial Intelligence (AI), Internet of Things (IoT), and Robotics is fundamentally transforming the design and operation of intelligent autonomous systems. Each of these domains independently enables critical capabilities: AI contributes data-driven reasoning and decision-making, IoT provides pervasive sensing and communication, and robotics delivers physical interaction with the environment. However, the combined potential of these technologies lies not in their individual functionalities but in their \emph{system-level integration}, wherein distributed intelligent agents can collectively sense, reason, and act in complex, dynamic, and partially observable environments.

Applications of such integrated systems are rapidly expanding across domains, including smart manufacturing~\cite{ji2019iot}, precision agriculture~\cite{romeo2020internet}, disaster response~\cite{afanasyev2019towards}, healthcare~\cite{pradhan2021iot}, and autonomous mobility~\cite{khalid2021internet}. These systems are expected to operate under real-time constraints, interact with uncertain environments, and adapt to evolving goals, necessitating seamless coordination between distributed sensing (IoT), machine reasoning (AI), and embodied control (Robotics). Achieving this coordination requires explicit architectural mechanisms for knowledge sharing, semantic integration, and decision-making across heterogeneous agents.

Despite these advances, most existing deployments remain siloed, combining only two out of the three technological pillars. For instance, the AIoT paradigm~\cite{siam2024aiot} focuses on intelligent sensing and analytics but largely omits actuation and embodied interaction. Similarly, the Internet of Robotic Things (IoRT)~\cite{afanasyev2019towards, khalid2021internet} emphasizes connectivity and robotic coordination but treats AI as a peripheral or abstract component. As a result, current systems lack the architectural consistency, modular scalability, and semantic integration required to support robust distributed reasoning and collective autonomy across domains.

\begin{figure}[!ht]
\centering
\resizebox{0.7\linewidth}{!}{%
\begin{tikzpicture}[
  node distance=0.8cm and 1.2cm,
  every node/.style={align=center, font=\small},
  main/.style={rectangle, draw, rounded corners, thick, fill=gray!15, minimum width=3.5cm, minimum height=1.2cm},
  section/.style={rectangle, draw, thick, fill=blue!10, minimum width=4.2cm, minimum height=1cm},
  arrow/.style={->, thick}
]


\node[section] (related) {Section II:\\Related Works};
\node[section, below=of related] (foundations) {Section III:\\Foundational Definitions \& Taxonomy};
\node[section, below=of foundations] (architecture) {Section IV:\\Architectural Patterns};
\node[section, below=of architecture] (applications) {Section V:\\Application Pillars};
\node[section, below=of applications] (distribution) {Section VI:\\Intelligence Distribution Models};
\node[section, below=of distribution] (challenges) {Section VII:\\Challenges \& Future Directions};
\node[section, below=of challenges] (conclusion) {Section VIII:\\Conclusion};

\draw[arrow] (related) -- (foundations);
\draw[arrow] (foundations) -- (architecture);
\draw[arrow] (architecture) -- (applications);
\draw[arrow] (applications) -- (distribution);
\draw[arrow] (distribution) -- (challenges);
\draw[arrow] (challenges) -- (conclusion);

\end{tikzpicture}
}
\caption{Paper structure and logical flow of sections. Each component builds toward a unified understanding of AI–IoT–Robotics integration.}
\label{fig:paper_structure}
\end{figure}
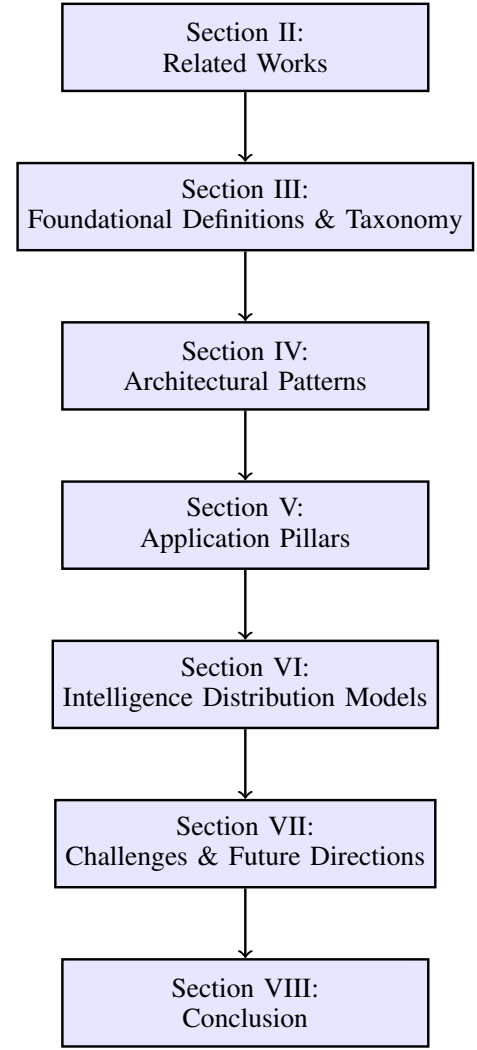

Moreover, there exists a notable absence of unified frameworks that explicitly model the feedback loops among perception, cognition, and action within heterogeneous distributed systems. The lack of standardized architectures impedes interoperability, while unresolved challenges in data fusion, real-time learning, and system-level security continue to hinder deployment at scale~\cite{zafir2024security}. These issues are further exacerbated in scenarios that require dynamic collaboration among multiple agents and edge intelligence under resource constraints, where reasoning must be distributed rather than centralized.

In this paper, we provide a comprehensive survey of the integration of Artificial Intelligence, Internet of Things, and Robotics, emphasizing unified frameworks, architectural strategies, and cross-domain challenges. Our objective is to consolidate existing approaches while identifying the emerging trends that are shaping the evolution of intelligent autonomous systems. Beyond technical synthesis, we argue that research is converging toward a new paradigm of Connected Robotics, in which robots operate as interconnected nodes within global cyber-physical ecosystems, and toward Physical AI, where intelligence is embodied through adaptive, sensorimotor interaction with the real world. This survey aims to serve both as a reference for current integration methodologies and as a roadmap for advancing toward these next-generation paradigms of autonomy. Future research must move beyond integration toward formal models of distributed reasoning and evaluation methodologies for collective intelligence in connected robotic systems. The overall structure of the paper is illustrated in Fig. \ref{fig:paper_structure}.

\section{Limitations of Prior Integration Studies and Need for a Unified Perspective}
\label{sec:prior_limits}

The convergence of AI, IoT, and Robotics has led to diverse survey literature exploring their pairwise integration. However, very few studies address all three domains simultaneously. Table~\ref{tab:related_work_summary} summarizes recent surveys: while each makes significant contributions within its scope, they often omit the broader system capabilities that arise only from full triadic integration of AI, IoT, and Robotics. The following discusses the limitations of each partial combination and introduces real-world challenges that require the combined strengths of all three.

\textbf{Robotics + IoT (without AI):} Works such as Ji et al.~\cite{ji2019iot} and Afanasyev et al.~\cite{afanasyev2019towards} emphasize architectures for the Internet of Robotic Things (IoRT). These systems improve remote operation and sensor fusion, but without AI, they lack learning or predictive control. Romeo et al.~\cite{romeo2020internet} categorize IoRT use cases across domains like smart factories and agriculture but do not address how systems adapt to changing environments. AI is necessary to add such adaptability. Tasks like predictive maintenance or anomaly detection in smart manufacturing demand cognitive intelligence that Robotics + IoT systems alone cannot provide.

\textbf{IoT + AI (without Robotics):} AIoT systems, such as those reviewed by Siam et al.~\cite{siam2024aiot}, excel in sensing and analytics. Healthcare applications~\cite{Bout2025, karar2022survey} demonstrate real-time monitoring and prediction, but without robots, they cannot physically assist patients. For example, predicting a fall or medication need does not close the autonomy loop without actuation. Robotic agents are required to transform these decisions into real-world interventions.

\textbf{AI+Robotics (without IoT):} Robotic systems with embedded AI excel in perception, planning, and adaptive control. However, they remain context-poor when disconnected from distributed sensor data. Dennison et al.~\cite{Dennison2025} show that hydroponic farming with multiple AI-driven robots connected via IoT sensors outperforms isolated robotics setups. Similarly, Kim et al.~\cite{kim2024survey} review Large Language Model (LLM)-enabled robotics and emphasize the requirement of IoT-generated environmental data for semantic reasoning and task execution. Together, these examples strongly support that AI and Robotics alone are insufficient without connectivity and shared situational awareness provided by IoT.

\textbf{Partial Coverage and Security-Focused Studies:} Zafir et al.~\cite{zafir2024security} and Khalid~\cite{khalid2021internet} mention all three domains but focus on security and lack operational implementations. They highlight architectural layers and risks but do not offer real-time, intelligent, and embodied integration across AI, IoT, and Robotics.

\textbf{Emerging Trends Highlight Full Integration:} The application of LLMs in robotics is not hypothetical, it is already being deployed with IoT support. Kim et al.~\cite{kim2024survey} categorize real-world systems where AI-driven reasoning, IoT-sensed context, and robotic embodiment converge for natural language understanding, adaptive navigation, and human–robot interaction. These systems demonstrate that only full AI + IoT + Robotics fusion enables semantic, interactive, and physically grounded behavior. A recent overview of AI-driven IoT applied to robotics is presented in \cite{restpublisher2025aidriveneiot}, offering a high-level discussion of emerging trends and challenges. However, it does not model the complete AI–IoT–Robotics loop, nor does it introduce taxonomies, unified architectures, or intelligence distribution analyses. Our work addresses these gaps by providing a structured system-level perspective, detailed architectural models, and a comprehensive integration survey spanning perception, cognition, and action.

These examples demonstrate that many real-world applications cannot be fulfilled using just two technologies. Our unified framework addresses this gap by embedding reasoning, connectivity, and embodiment into one cohesive system.

\begin{table*}[ht]
\centering
\caption{Comparison of Key Survey Works on AI, IoT, and Robotics Integration}
\label{tab:related_work_summary}
\begin{tabular}{|p{3cm}|c|c|c|p{10cm}|}
\hline
\textbf{Citation} & \textbf{AI} & \textbf{IoT} & \textbf{Robotics} & \textbf{Limitations} \\
\hline
Afanasyev et al.~\cite{afanasyev2019towards} & \xmark & \cmark & \cmark & Focuses on IoRT architectures; lacks AI analysis \\
\hline
Ji et al.~\cite{ji2019iot} & \xmark & \cmark & \cmark & Emphasizes IoT–robotics in manufacturing; excludes AI-driven capabilities \\
\hline
Romeo et al.~\cite{romeo2020internet} & \xmark & \cmark & \cmark & Reviews domain-specific IoRT use cases; limited focus on learning-enabled robotics \\
\hline
Khalid et al.~\cite{khalid2021internet} & \cmark & \cmark & \cmark & Discusses AI conceptually; lacks unified architectural model \\
\hline
Pradhan et al.~\cite{pradhan2021iot} & \xmark & \cmark & \cmark & Focused on healthcare; does not consider autonomous reasoning or AI control \\
\hline
Siam et al.~\cite{siam2024aiot} & \cmark & \cmark & \xmark & Covers AIoT extensively; robotics mentioned only as endpoint applications \\
\hline
Zafir et al.~\cite{zafir2024security} & \cmark & \cmark & \cmark & Focuses on security; lacks real-time AI-based adaptation in robotics \\
\hline
Alam et al.~\cite{restpublisher2025aidriveneiot} & \cmark & \cmark & \cmark & 
Provides a broad high-level overview of AI and IoT in robotics; lacks unified taxonomy, system-level architecture, and triadic AI–IoT–Robotics integration. \\
\hline

\end{tabular}
\end{table*}

\section{Foundational Definitions \& Taxonomy}

The development of autonomous systems that combine AI, IoT, and Robotics requires a precise understanding of the roles, capabilities, and integration points of each technology. In this section, we define each component within the context of system-level intelligence and introduce a layered taxonomy that classifies their integration.

\subsection{Definitions}

\textbf{Artificial Intelligence (AI)} is a field of computer science focused on developing algorithms and systems that can perceive, reason, learn, and make decisions~\cite{russell2016artificial}. In the context of integrated systems, AI enables interpretation of sensory input, pattern recognition, adaptive behavior, and decision-making under uncertainty. It enhances both perception and planning capabilities in autonomous and distributed environments~\cite{chen2021deep, zhang2020empowering_aiot}.

\textbf{Internet of Things (IoT)} refers to the network of interconnected devices embedded with sensors, actuators, and communication interfaces that collect and exchange data in real-time~\cite{atzori2010internet, madakam2015internet, liu2020survey}. In multi-agent or cyber-physical systems, IoT provides distributed situational awareness, remote monitoring, and infrastructure for coordination and data fusion across spatially separated components~\cite{zhou2019edge, al2015internet}.

\textbf{Robotics} involves the design and deployment of systems capable of perceiving their surroundings, making decisions, and physically interacting with the environment through actuation and control~\cite{siciliano2016springer, kormushev2013reinforcement}. Robotics offers the physical embodiment necessary for autonomous tasks, including manipulation, mobility, and real-time adaptation to dynamic settings.

Each component - AI, IoT, and Robotics - plays a distinct but interdependent role in realizing intelligent, responsive, and connected systems capable of operating in complex real-world scenarios.

\subsection{Integration Perspective}

Although each technology independently enables a subset of intelligent behavior, their synergy creates systems with higher degrees of autonomy and adaptability. For example, AI enables predictive control, IoT provides remote monitoring and distributed coordination, while robotics delivers real-time interaction with the environment~\cite{ji2019iot, khalid2021internet, siam2024aiot}. When jointly deployed, these technologies form a feedback-driven, closed-loop intelligent system capable of operating at the edge and reacting to spatiotemporal dynamics~\cite{krejci2025_internet_of_robotic_things}. Throughout this survey, the AI layer encompasses a wide range of techniques, including perception models, reinforcement learning, planning algorithms, symbolic reasoning, and domain-specific analytics. The discussion of SLMs and LLMs in later sections serves only to highlight an emerging trend in distributed AI, rather than to represent the entire AI domain.

\subsection{Taxonomy of AI--IoT--Robotic Systems}

\begin{figure}[ht]
\centering
\resizebox{0.9\linewidth}{!}{%
\begin{tikzpicture}[
  font=\sffamily\small,
  box/.style={draw, rounded corners, thick, align=center, minimum width=3.6cm, minimum height=1.2cm},
  arrow/.style={->, thick}
]

\node[box, fill=gray!10]   (ai)   at (0,3.0)  {AI Cognition \\ (Perception, Planning, Learning)};
\node[box, fill=blue!10]   (iot)  at (-3,0.0) {IoT Perception \\ (Distributed Sensors \& Devices)};
\node[box, fill=orange!10] (robot)at (3,0.0)  {Robotics Action \\ (Mobile \& Manipulator Robots)};

\draw[arrow] (iot.north east)  -- node[left,  xshift=-1mm, yshift=1mm] {\footnotesize Data Streams} (ai);
\draw[arrow] (ai)   -- node[right, xshift=0mm,  yshift=1mm] {\footnotesize Commands \& Policies} (robot.north west);
\draw[arrow] (robot)-- node[below, yshift=-6mm] {\footnotesize Environment Interaction \& Feedback} (iot);

\draw[arrow, dashed, bend left=20]  (iot)  to node[above, xshift=-10mm, yshift=1mm] {\footnotesize Context Updates} (ai.west);
\draw[arrow, dashed, bend right=20]  (robot)to node[above, xshift=8mm, yshift=1mm] {\footnotesize State Reports} (ai.east);

\node at (0,4.0) {\textbf{Unified AI--IoT--Robotics Integration Loop}};

\end{tikzpicture}
}
\caption{Conceptual closed-loop integration between IoT-based perception, AI cognition, and robotic action. IoT devices provide distributed sensing, AI models transform data into decisions, and robots act in the environment while feeding back new observations.}
\label{fig:ai_iot_robotics_loop}
\end{figure}
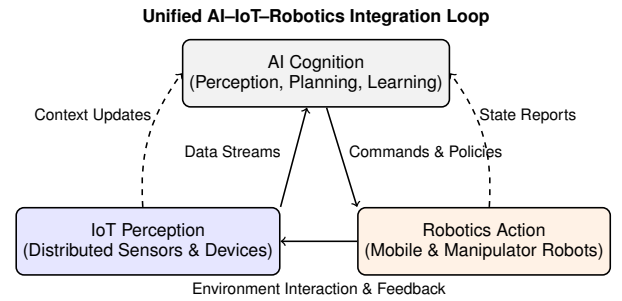

We propose a three-layer taxonomy to classify the interaction and integration of AI, IoT, and Robotics, as illustrated in Figure \ref{fig:ai_iot_robotics_loop}.

\begin{enumerate}
    \item \textbf{Perception Layer (IoT-centric):} Comprises distributed sensors and IoT devices responsible for environmental monitoring, localization, and data transmission~\cite{madakam2015internet, zhou2019edge}. Data acquired here includes physical measurements (e.g., temperature, location, speed) and high-level observations (e.g., images, speech). A practical example of this is MoCArU, a portable, low-cost robot localization system using IoT markers and wireless communication \cite{DBLP:conf/smc/AssabumrungratB23, DBLP:conf/sii/BarrosBAKOKOT25}.

    \item \textbf{Cognition Layer (AI-centric):} Hosts computational models, including machine learning, deep learning, and symbolic reasoning techniques~\cite{pan2020edge, zhang2020empowering_aiot}. This layer processes raw data into actionable insights, such as anomaly detection~\cite{chen2021deep} or task planning~\cite{kormushev2013reinforcement}.
    
    \item \textbf{Action Layer (Robotics-centric):} Implements the physical interface through which the system acts, including mobile and manipulator robots, embedded controllers, and motorized platforms~\cite{siciliano2016springer}. Commands from the cognition layer are translated into executable motor actions, forming a closed control loop.
\end{enumerate}

This taxonomy highlights the hierarchical and functional separation among sensing, reasoning, and actuation components, while emphasizing their bidirectional interactions. Notably, AI may be embedded at the edge (e.g., within IoT nodes), centralized in the cloud, or distributed across robotic platforms~\cite{zhu2020toward_intelligent_edge}. Similarly, IoT protocols enable inter-robot coordination in multi-agent settings~\cite{al2015internet}, and robotic agents may act as mobile IoT nodes~\cite{krejci2025_internet_of_robotic_things}.

While some recent works have proposed architectural perspectives on IoRT or AIoT systems~\cite{khalid2021internet, siam2024aiot}, none offer a unified taxonomy that models the structural and functional relationships across all three domains. Our proposed taxonomy fills this gap by capturing the full system stack and its flow of information, enabling consistent analysis and implementation across disciplines.

\section{Architecture Proposals}

Designing scalable, responsive, and intelligent systems that integrate AI, IoT, and Robotics requires a robust architectural foundation. In this section, we review common architectural patterns employed in the literature and propose a unified system design to support dynamic interaction between perception, cognition, and actuation. We also highlight prevailing integration challenges that impact real-world deployment.

\subsection{Common Architectural Patterns}

\subsubsection{Edge--Fog--Cloud Hybrid}

One widely adopted pattern is the hierarchical \textit{edge--fog--cloud} architecture~\cite{zhou2019edge, zhu2020toward_intelligent_edge, pan2020edge}. In this design, data acquisition occurs at the \emph{edge layer}, typically composed of IoT sensors and embedded processors close to the physical environment. Preliminary processing (e.g., filtering, event detection) may be executed on these nodes, minimizing latency and communication overhead~\cite{al2015internet, krejci2025_internet_of_robotic_things}.

The \emph{fog layer} acts as an intermediate computation tier. It aggregates data from multiple edge nodes, performs context-aware analytics, and may control multiple robots in a local network~\cite{bonomi2012fog, mach2017mobile}. This layer often hosts reinforcement learning models, task allocation policies, or collaborative planning modules~\cite{abdel2018fog}. The \emph{cloud layer} is responsible for high-complexity operations, such as deep learning inference, cross-domain reasoning, and long-term memory storage~\cite{zhang2020empowering_aiot}.

This architecture enables system scalability and supports both low-latency control and high-throughput data processing. However, effective operation depends on (i) \emph{middleware choices} for reliable edge--edge and edge--cloud communication in distributed ROS~2 deployments~\cite{zhang2024ros2_middleware}, and (ii) \emph{latency-aware placement and scheduling} of dependent services across edge and cloud~\cite{sun2025latencyaware}. In interactive workloads, QoS-aware offloading and resource allocation further improve responsiveness, but remain sensitive to connectivity variability and model mismatch~\cite{hao2025qos_ar_offloading}.

\subsubsection{Peer-to-Peer Swarm Intelligence}

Swarm robotics and decentralized multi-agent systems adopt a \textit{peer-to-peer} architecture in which each robotic node communicates directly with its neighbors~\cite{brambilla2013swarm, sabattini2013distributed_global_connectivity, arafat2021swarm}. Such designs are inspired by biological collectives and are advantageous for tasks like area coverage, search and rescue, or environmental monitoring.

In swarm-based systems, AI components (e.g., local planners, classifiers, and allocation heuristics) are distributed across agents. Each robot acts as both an actuator and a sensor node, sharing information through direct inter-agent protocols and local neighborhood consensus~\cite{brambilla2013swarm, sabattini2013distributed_global_connectivity, arafat2021swarm}. This architecture improves robustness to single points of failure and supports graceful degradation when connectivity to infrastructure is intermittent.

The primary limitation is that local-only computation and communication impose strict constraints on model size, bandwidth, and real-time reliability. Practical deployments therefore rely on efficient on-device inference (including memory-efficient long-context inference where language-enabled reasoning is used)~\cite{chen2025edgeinfinite}, model compression for constrained devices~\cite{zhu2024llm_compression_survey}, and selective information exchange policies to control communication cost while preserving coordination performance.

\subsection{LLM-SLM-Enabled AIoT–Robotics Architecture}

\begin{figure}[ht]
\centering
\resizebox{0.9\linewidth}{!}{%
\begin{tikzpicture}[
  font=\sffamily\small,
  box/.style={draw, rounded corners, thick, align=center, minimum width=3.2cm, minimum height=1.2cm},
  edgecloud/.style={draw=black, thick, align=center, minimum width=3cm, minimum height=1cm},
  io/.style={draw, thick, dashed, align=center, minimum width=2.6cm, minimum height=1cm},
  arrow/.style={->, thick}
]

\node[box, fill=gray!10] (cloud) at (0,4.5) {Cloud AI \\ (LLM Training \& Global Reasoning)};
\node[box, fill=gray!10] (fog) at (-4,2.5) {Fog Node \\ (Task Coordination)};
\node[box, fill=gray!10] (edge) at (4,2.5) {Edge AI \\ (SLMs, On-device Inference)};

\node[io, fill=blue!10] (iot1) at (-3.5,0) {IoT Sensor \\ (Gas, Temp)};
\node[io, fill=blue!10] (iot2) at (3.5,0) {IoT Actuator \\ (Valves, Switches)};

\node[box, fill=orange!10] (robot) at (0,-2.) {Robot \\ Perception $\leftrightarrow$ Action};

\draw[arrow] (cloud) -- (fog);
\draw[arrow] (cloud) -- (edge);

\draw[arrow] (fog) -- (iot1);
\draw[arrow] (fog) -- (robot);
\draw[arrow] (edge) -- (iot2);
\draw[arrow] (edge) -- (robot);

\draw[arrow, dashed] (iot1) -- (robot);
\draw[arrow, dashed] (robot) -- (iot2);

\draw[arrow, bend right=20] (robot.north) to node[left, xshift=-2mm, yshift=2mm, rotate=0]{\footnotesize Feedback Loop} (fog.east);
\draw[arrow, bend left=20] (robot.north) to node[right, xshift=2mm, yshift=2mm, rotate=0]{\footnotesize Decision Update} (edge.west);

\node at (0,5.5) {\textbf{SLM–LLM Integrated Architecture for AIoT–Robotic Systems}};

\end{tikzpicture}
}
\caption{Layered integration of SLMs at the edge and LLMs in the cloud for enabling perception, reasoning, and actuation in AIoT–robotic systems.}
\label{fig:unified_system_view}
\end{figure}
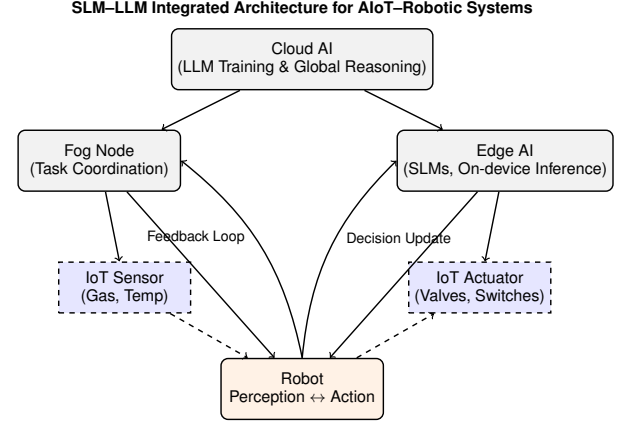

To illustrate one emerging trend within the AI domain, we describe how recent advances in Small and Large Language Models may operate across the edge--fog--cloud hierarchy. Recent progress in on-device language inference has improved feasibility under tight memory budgets~\cite{chen2025edgeinfinite}, while systematic compression approaches support practical deployment of language models on constrained platforms~\cite{zhu2024llm_compression_survey}. In parallel, system responsiveness depends on communication and orchestration mechanisms, including ROS~2 middleware selection for distributed deployments~\cite{zhang2024ros2_middleware} and latency-aware scheduling across IoT--edge--cloud infrastructures~\cite{sun2025latencyaware}. Learning-based optimization has also been explored in communication-centric IoT infrastructures, where reinforcement learning methods such as proximal policy optimization are applied to jointly optimize energy efficiency and transmission performance under dynamic network conditions~\cite{2023_Zhang}.
We emphasize that SLMs and LLMs are used here as illustrative examples of AI components that can support distributed reasoning alongside perception models, reinforcement learning, planning, and control~\cite{wang2025empowering}. This layered integration enables robust perception, adaptive reasoning, and real-time actuation, especially in edge environments constrained by bandwidth and latency.

\begin{enumerate}
    \item \textbf{Perception Layer:} Comprises distributed IoT sensors and robot-mounted cameras or LiDARs, capturing environmental data and user inputs. This data feeds both SLMs at the edge and LLMs in the cloud via encoding or summarization~\cite{zhu2020toward_intelligent_edge, zhou2019edge}.

    \item \textbf{Reasoning Layer:} 
    Incorporates edge intelligence for low-latency inference on local devices~\cite{zhou2019edge}, including memory-efficient long-context processing when language-based reasoning is required~\cite{chen2025edgeinfinite}. For global decision support, intent recognition, and high-level task reasoning, cloud-side models and supporting services may be used~\cite{kim2024survey}. Hybrid learning occurs through knowledge distillation or selective prompting across both, with deployability improved by modern compression pipelines~\cite{zhu2024llm_compression_survey}.

    \item \textbf{Actuation Layer:} Robots and actuators execute high-level plans from reasoning modules. Feedback (e.g., failed actions) is captured and fed back to upstream models for error correction or task reallocation~\cite{siciliano2016springer}.

    \item \textbf{Adaptation Layer:} Enables context-aware adjustment via local policy refinement and global retraining. Edge-fog-cloud communication ensures model updates and robustness to dynamic environments~\cite{wang2025empowering, hewitt2025health}. 
\end{enumerate}

This hybrid architecture reflects the current trend of decentralizing advanced inference capabilities through edge-deployable models, enabling AI--IoT--Robotic integration that is scalable, responsive, and privacy-aware~\cite{wang2025empowering}. At the systems level, robustness hinges on communication substrates and orchestration strategies (e.g., ROS~2 middleware choices and cross-layer scheduling)~\cite{zhang2024ros2_middleware, sun2025latencyaware}. It is important to note that SLMs and LLMs constitute only one subset of the broader AI landscape considered in this architecture. The framework accommodates diverse AI models, vision-based classifiers, multimodal fusion networks, reinforcement learning policies, and classical control algorithms, which may operate independently or in combination with language-model-based reasoning.

\subsection{Integration Challenges}

Several challenges impede the seamless integration of AI, IoT, and Robotics in unified systems:

\textbf{1) Latency and Real-Time Constraints:} AI-based perception and planning modules require low-latency data streams, which may be compromised by network delays, particularly when cloud services are used~\cite{zhou2019edge, zhang2020empowering_aiot}.

\textbf{2) Synchronization and Coordination:} Multi-agent robotic systems often rely on synchronized perception and control across distributed nodes. Ensuring temporal consistency and data coherence remains a critical hurdle~\cite{sabattini2013distributed_global_connectivity, xu2020edge}.

\textbf{3) Heterogeneity of Devices:} IoT networks typically consist of heterogeneous devices with varying computational power, operating systems, and communication protocols. This diversity complicates software deployment and model portability~\cite{liu2020survey}.

\textbf{4) Bandwidth and Communication Load:} High-resolution sensors and continuous video/audio streams demand significant bandwidth. Efficient data compression and prioritization strategies are required to maintain responsiveness~\cite{mach2017mobile}. Emerging IoT protocols such as CoAP can mitigate communication overhead and energy consumption; for example, Sarkar et.al. \cite{Sarkar2025} demonstrate that CoAP enables reliable, low-latency communication for mobile robot control while reducing packet overhead compared to HTTP or MQTT.

\textbf{5) Compute Limitations at the Edge:} Embedding complex AI models into edge and mobile robotic devices is constrained by limited hardware resources and energy budgets~\cite{pan2020edge, zhu2020toward_intelligent_edge}.

Addressing these integration challenges is fundamental to unlocking the full potential of AI--IoT--Robotic systems. Future research must explore adaptive architectures, standardized middleware, and learning-based scheduling mechanisms to balance performance and resource usage across the system.

\section{Application Pillars}

The convergence of AI, IoT, and Robotics enables transformative capabilities across multiple domains. In this section, we highlight five representative application areas, disaster response, autonomous logistics, smart factories, ambient assistive living, and agriculture where integrated intelligent systems have demonstrated both promise and challenges. In the following subsections, each application domain is described in terms of the concrete capabilities enabled by AI–IoT–Robotic integration, the specific problems addressed, the remaining technical challenges, and the typical physical/system configurations used in practice.

\subsection{Disaster Response and Hazard Mitigation}

Disaster-response systems that integrate AI, IoT, and robotics typically operate as multi-layered infrastructures composed of fixed and deployable sensors, autonomous robots, and edge or fog computing nodes. In real deployments, IoT devices such as structural strain gauges, radiation dosimeters, gas detectors, and thermal cameras are positioned throughout hazardous environments, including damaged buildings, tunnels, and nuclear facilities, to provide continuous situational measurements~\cite{yokokohji2021nuclear_accidents, okuzumi2024irid_fukushima_robotics, tadokoro2019rescue}. These sensors form the environmental backbone of the system, enabling remote teams and robotic platforms to monitor dynamic hazards such as aftershocks, gas leaks, or temperature spikes.

Mobile robots, ranging from tracked ground vehicles to UAVs and inspection robots for severe environments, are deployed to enter areas unsafe for humans, where they acquire high-resolution visual, thermal, and LiDAR data and support risk-aware decision-making~\cite{drew2021multiagent_sar, yokokohji2021nuclear_accidents, kojima2022wrs2020_plant_inspection}. Their on-board AI modules perform tasks such as collaborative mapping and situation reconstruction~\cite{chatziparaschis2020uav_ugv_mapping}, anomaly detection in sensor and LiDAR streams~\cite{DBLP:conf/icar/HattoriBKOOIST23}, and autonomous navigation in cluttered or partially collapsed structures. These capabilities are complemented by edge/fog computing nodes placed at the periphery of the disaster site, which aggregate multimodal data from IoT sensors and robots to enable low-latency reasoning and mission-level decision support under bandwidth constraints~\cite{singh2023fog_edge_ai_robotics, tahir2025edge_robotics_survey}.

The integration of these components results in practical workflows that extend beyond simple sensing--reasoning--acting dynamics. For example, UAV--UGV teams can collaboratively explore and map disaster sites, generating task-relevant 3D representations that support structural assessment and victim search while limiting human exposure~\cite{chatziparaschis2020uav_ugv_mapping, measurement2024uav_pointcloud_damage}. AI algorithms fuse heterogeneous data streams to generate updated hazard maps, recommend safe robot trajectories, and detect anomalies such as unexpected heat sources or structural deformation. Such coordinated operation has been instrumental in real-world response efforts, particularly in nuclear-accident contexts where remote sensing and robotic inspection are required inside high-radiation areas~\cite{yokokohji2021nuclear_accidents, okuzumi2024irid_fukushima_robotics}.

Despite these advances, several barriers remain. Reliable communication is difficult to maintain in underground or metal-reinforced environments, causing frequent data loss or asynchrony across IoT devices and robots. Heterogeneous sensor calibration, multimodal data fusion, and real-time AI inference under strict latency and energy constraints remain open challenges. Furthermore, unpredictable terrain and rapidly evolving hazards demand robust autonomy and standardized evaluation methods that current systems only partially achieve~\cite{tadokoro2019rescue, kojima2022wrs2020_plant_inspection, drew2021multiagent_sar}.

\subsection{Autonomous Logistics and Transportation}

AI--IoT--Robotic systems in logistics are typically deployed as multi-layered
infrastructures combining IoT-based asset tracking, autonomous mobile robot
fleets, and AI-driven coordination platforms. In practical warehouse and
distribution deployments, IoT devices such as RFID tags, UWB anchors, and
barcode or QR markers provide continuous identification and localization of
parcels, racks, and pallets throughout the facility, enabling end-to-end
visibility across transportation, warehousing, and handling operations%
~\cite{song2020iot_smart_logistics_survey, golpira2021_logistics_iot_review}.
These data streams are aggregated at edge or fog computing nodes, where AI
models perform real-time state estimation, task allocation, and fleet-level
scheduling under latency and bandwidth constraints~\cite{zhou2019edge}.

Mobile robots, including AGVs and AMRs, are equipped with cameras, LiDAR
sensors, and (when required) tag readers, enabling autonomous navigation in
cluttered aisles and safe material transport with minimal human intervention.
Core challenges and solution families for AGV system design and control include
guide-path and layout design, routing, dispatching and scheduling, deadlock
avoidance, and battery management~\cite{le_anh2006_agv_review}. At the fleet level, coordination services at the fog layer balance workload, reduce
congestion, and mitigate energy and throughput trade-offs~\cite{zhou2019edge}.

A typical workflow begins when real-time logistics signals, such as shipment
status, inventory changes, and workstation requests, are incorporated into a
coordination layer that updates system state and triggers (re)assignment of
transportation or handling tasks.
~\cite{firdausiyah2019city_logistics_mas_adp}. Robots continuously report state (location, progress, queue times) back to the coordination layer, enabling dynamic replanning in response to human activity,
unexpected obstacles, or traffic congestion; edge/fog processing further
ensures low-latency updates and reduces dependence on cloud connectivity%
~\cite{zhou2019edge}.

Despite these advances, several integration challenges remain. Dynamic obstacle
handling in densely populated facilities requires robust multi-sensor fusion
and reliable human--robot interaction strategies. Interoperability between
legacy industrial systems and new IoT--robotic platforms remains limited,
complicating deployment in existing facilities. Moreover, the exchange of
operational and asset data across large-scale networks introduces security and
privacy vulnerabilities that must be carefully managed%
~\cite{song2020iot_smart_logistics_survey,golpira2021_logistics_iot_review}.

\subsection{Smart Factories and Industrial Automation}

The vision of Industry~4.0 hinges on the fusion of AI, IoT, and Robotics in manufacturing environments~\cite{lu2017industry, ji2019iot, wang2016towards}. Smart factories deploy dense sensor and actuator networks to monitor machine status, process variables, product quality, and human--robot interaction, feeding data into AI systems for quality inspection, predictive maintenance, and adaptive production~\cite{zheng2018smart, cie2020_pdm_slr}. Increasingly, these pipelines are being implemented as cyber--physical systems that combine shop-floor sensing with edge/fog computing to reduce latency and bandwidth while enabling near-real-time control loops~\cite{ins2021_edge_industrial_internet, farooq2023_iiot_smart_industry}.

Digital twins have become a core enabler for closed-loop optimization in smart factories by linking live IIoT telemetry to simulation and analytics for monitoring, what-if evaluation, and continual improvement of production plans and robot behaviors~\cite{huang2021_ai_digital_twin_survey}. In parallel, collaborative robots (cobots) are moving beyond pre-programmed routines toward learning-based interaction, where imitation learning and reinforcement learning support task adaptation and safer human--robot collaboration on assembly lines~\cite{villani2018survey, rcim2022_hrc_robot_learning_survey}. AI models further assist in anomaly detection, fault diagnosis, and operational optimization from high-frequency sensor streams~\cite{kang2016smart, cie2020_pdm_slr}.

Recent work has also demonstrated how cooperative multi-robot pickup and delivery systems can enable customized production lines, where heterogeneous robots coordinate task allocation and scheduling under dynamic manufacturing conditions~\cite{9926509, DBLP:conf/case/BezerraOKAGKOKT23}. These studies illustrate how AI-based coordination layers can be tightly coupled with IoT-enabled production infrastructure, enabling closed-loop adaptation between sensing, task allocation, and physical execution in smart factories. Beyond the factory floor, the same coordination primitives support flexible material handling and intralogistics, where fleet-level dispatching must account for congestion, safety, and energy constraints.

Despite these advances, several integration barriers remain. Interoperability and data standardization across heterogeneous machines, vendors, and software stacks continue to limit plug-and-play deployment. Synchronization between digital twins, edge analytics, and robot controllers under stringent latency constraints is still challenging, especially when connectivity is intermittent or when safety-certified control must coexist with learning-based modules~\cite{ins2021_edge_industrial_internet, huang2021_ai_digital_twin_survey}. Finally, scalable evaluation benchmarks for end-to-end learning, maintenance, and multi-robot coordination in real factories remain an important open direction~\cite{farooq2023_iiot_smart_industry, cie2020_pdm_slr}.

\subsection{Ambient Assistive Living and Healthcare Robotics}

AI--IoT--Robotic systems are increasingly applied to support elderly care, rehabilitation, and health monitoring in ambient assistive living environments~\cite{pradhan2021iot, price2017ai_healthcare_legal, ando2021iort_aal_patterns}. Wearable and ambient IoT sensors capture physiological and behavioral data, which AI systems use to detect anomalies, assess risk, and trigger interventions~\cite{li2017iothealth}.

Service robots perform tasks such as medication delivery, fall detection response, or social interaction. Cognitive architectures enable these robots to interpret user intent, adapt to individual preferences, and provide personalized assistance~\cite{kostavelis2015semantic}. Integration with smart home devices allows for seamless interaction and environment adaptation.

Recent literature has begun to explore the integration of emotional AI into healthcare robotics, aiming to provide empathy‑driven responses alongside physiological monitoring and mobility assistance.  Bout et.al. \cite{Bout2025} review this emerging field, emphasizing both its promise for patient‑centred care and the ethical challenges posed by bias and privacy concerns.  Incorporating such perspectives broadens the scope of assistive‑robotics applications and underscores the importance of ethically aligned AI.

Privacy concerns, limited on-device processing, and interoperability among health standards and robotic platforms are critical barriers that must be addressed.

\subsection{Smart Agriculture and Aquaculture}

Recent advances in smart farming illustrate how tightly integrated AI, IoT, and robotics can enhance productivity, resource efficiency, and operational robustness across both controlled and open-field settings. In controlled-environment agriculture (CEA), dense IoT sensing (e.g., pH/EC, temperature, humidity, dissolved oxygen, lighting) coupled with automation enables closed-loop nutrient, climate, and water management for large-scale hydroponics and aquaponics~\cite{Dennison2025}. Beyond automation, deep learning has become central to CEA for microclimate prediction, stress and disease detection, and yield-related monitoring under variable conditions~\cite{Ojo2022CEAReview}. 

In open-field precision agriculture, heterogeneous robotic teams (e.g., UAV--UGV cooperation) use distributed sensing and planning to improve coverage and timeliness of monitoring and intervention, reducing labor demands while improving data quality for downstream AI analytics~\cite{tokekar2013sensor}. To meet real-time constraints and intermittent connectivity in rural deployments, recent work increasingly emphasizes edge/near-sensor inference and hierarchical processing pipelines that reduce bandwidth usage and latency while maintaining robust decision support~\cite{Sensors2025AIoTAgriculture}.

Similar AIoT principles are now widely adopted in aquaculture. Continuous water-quality sensing (e.g., pH, turbidity, ammonia, dissolved oxygen, temperature), combined with predictive analytics, supports early anomaly detection, feeding optimization, and risk-aware farm operation. In shrimp farming specifically, recent surveys highlight how AIoT-based monitoring and decision support can improve yield and resource efficiency while raising challenges in calibration, maintenance, and data reliability in harsh wet environments~\cite{DuyLe2025}. More broadly, AIoT-driven aquaculture has expanded to include fish behavior analysis, disease monitoring, and automated control loops for aeration and feeding~\cite{Huang2025AIoTAquaculture}.

Despite these advances, deployment barriers remain substantial, including high upfront investment, sensor drift and biofouling, energy constraints, and limited interoperability across vendors and farm management platforms. Energy management and coordination are particularly critical for long-duration agricultural robot operations. Carvalho et al.~\cite{DBLP:conf/larc/CarvalhoBRM24} demonstrate reinforcement-learning-based task allocation for multi-robot wireless charging in agricultural IoT networks, improving fleet uptime and reducing downtime. Complementary studies on agricultural robotic charging infrastructure further show the importance of practical power-delivery and charging-system engineering for sustained autonomy in the field~\cite{Bodian2025WirelessChargingAgriRobots}.

\section{Intelligence Distribution Models}

The deployment of intelligence in AI--IoT--Robotics systems can follow multiple paradigms depending on application requirements, network constraints, safety requirements, and autonomy levels. This section discusses centralized, decentralized, and federated intelligence distribution models and examines emerging roles of multi-agent learning, Large Language Models (LLMs), and Small Language Models (SLMs) in real-world robotics.

While Section IV describes architectural patterns such as the edge--fog--cloud hierarchy from a system-design perspective, the present section focuses specifically on how \emph{intelligence} is distributed across those layers. Accordingly, Section VI examines the placement and training/execution split of AI models (centralized, decentralized, federated) rather than the communication topology. Table~\ref{tab:intelligence_models} presents representative works across these paradigms, emphasizing practical deployment constraints (latency, connectivity, compute budgets, and system heterogeneity).

Beyond traditional cloud-centric and device-centric paradigms, recent networking research has investigated LLM-driven generative AI agents to support distributed reasoning and decision-making under bandwidth and latency constraints~\cite{zhang2024jsac_moeagents}. In AIoT--robotics settings, such cloud-centric coordination must be coupled with engineering mechanisms that determine where services run and how they communicate, such as latency-aware edge--cloud scheduling~\cite{sun2025latencyaware} and middleware choices for distributed ROS~2 edge-to-edge and edge-to-cloud deployments~\cite{zhang2024ros2_middleware}. Related work has also explored interactive retrieval-augmented mechanisms for semantic context retrieval and information management in next-generation networks~\cite{2024_Zhang}, which can complement distributed decision assistance in heterogeneous systems.

\begin{table*}[t!]
\centering
\caption{Survey of Intelligence Distribution Models: Advantages, Disadvantages, and Representative Contributions}
\label{tab:intelligence_models}
\begin{tabular}{|p{2.2cm}|p{2.2cm}|p{0.5cm}|p{2.2cm}|p{2.6cm}|p{2.6cm}|p{2.6cm}|}
\hline
\textbf{Citation} & \textbf{Distribution Model} & \textbf{Year} & \textbf{Target System} & \textbf{Advantages} & \textbf{Disadvantages} & \textbf{Representative Contribution} \\
\hline

\multicolumn{7}{|c|}{\textbf{Centralized / Cloud-Centric Intelligence}} \\
\hline
Zhang et al.~\cite{zhang2024jsac_moeagents} &
Centralized (Cloud AI) &
2024 &
Satellite communication networks &
Cloud-scale reasoning and coordination; supports complex decision pipelines with LLM agents &
Not validated for embodied robotics; depends on network reliability and centralized orchestration &
LLM-driven generative agents with MoE-based transmission optimization for communication-aware decision-making \\
\hline

Vermesan et al.~\cite{vermesan2020iort_platforms} &
Centralized (Cloud/Platform-centric IoRT) &
2020 &
IoRT platforms and connectivity &
Architectural basis for large-scale IoRT integration, connectivity, and trust frameworks &
High-level architecture; limited quantitative evidence for real-time control loops &
Defines IoRT connectivity/platform concepts supporting cloud-centric orchestration of robotic things \\
\hline

\multicolumn{7}{|c|}{\textbf{Hybrid Collaboration (Edge--Cloud / Edge--Edge--Cloud)}} \\
\hline
Hao et al.~\cite{hao2025qos_ar_offloading} &
Offloaded (Cloud--Edge) &
2025 &
AR task offloading &
QoS-driven decomposition and placement across cloud-edge; practical for interactive workloads &
Assumes accurate performance models; sensitive to connectivity variability &
QoS-aware AR subtask offloading and resource allocation in cloud--edge collaboration environments \\
\hline

Sun et al.~\cite{sun2025latencyaware} &
Hybrid (IoT--Edge--Cloud scheduling) &
2025 &
Collaborative IoT-edge-cloud networks &
Explicitly optimizes latency satisfaction and load balancing under task dependencies; relevant to time-sensitive sensing pipelines &
Assumes reliable latency modeling and coordination; robustness under intermittent connectivity and real robotic sensing/actuation loops requires further validation &
Latency-aware scheduling for data-oriented service requests decomposed into dependent tasks across edge and cloud \\
\hline

Zhang et al.~\cite{zhang2024ros2_middleware} &
Hybrid (Edge--Edge / Edge--Cloud networking) &
2024 &
Distributed ROS~2 systems &
Directly relevant to robotics deployments; comparative middleware evidence for distributed setups &
Focuses on communication performance; does not solve higher-level autonomy distribution by itself &
Quantitative comparison of middleware choices for ROS~2 distribution across edge-to-edge and edge-to-cloud \\
\hline

Schaff and Walter~\cite{schaff2020residual_shared_autonomy} &
Human-in-the-loop (Shared autonomy) &
2020 &
Assistive robotics &
Balances local autonomy with human control; improves performance without full intent modeling &
Requires careful constraint design; depends on user interaction assumptions &
Residual policy learning enabling shared autonomy via minimal corrective actions under constraints \\
\hline

\multicolumn{7}{|c|}{\textbf{Decentralized Intelligence (On-device / Local AI)}} \\
\hline
Chen et al.~\cite{chen2025edgeinfinite} &
Decentralized (On-device LLM/SLM) &
2025 &
Edge language models &
Enables longer-context processing under tight memory budgets; improves responsiveness on-device &
Still limited by edge compute and latency/accuracy trade-offs for real-time robotics &
Memory-efficient infinite-context Transformer design targeted for edge devices \\
\hline

Zhu et al.~\cite{zhu2024llm_compression_survey} &
Decentralized (Model efficiency for on-device LLMs) &
2024 &
Compressed LLMs for constrained devices &
Strong credibility anchor; consolidates pruning/quantization/distillation for deployable LLMs &
Survey paper; robotics embodiment validation not the main focus &
Comprehensive survey of LLM compression techniques and evaluation practices for practical deployment \\
\hline

Pinto et al.~\cite{pinto2018asymmetric_rss} &
Decentralized (On-device RL policy execution) &
2018 &
Robot policy learning &
Real-time local control; robust to cloud dropouts &
Training is expensive; not designed for fleet-scale distribution by default &
Asymmetric actor--critic training leveraging privileged simulator state for improved vision-policy learning \\
\hline

\multicolumn{7}{|c|}{\textbf{Federated / Distributed Multi-Agent Learning}} \\
\hline
Foerster et al.~\cite{foerster2018coma} &
CTDE MARL (Centralized training + decentralized execution) &
2018 &
Cooperative multi-agent systems &
Classic, widely cited conceptual anchor for distributed team learning and credit assignment &
CTDE assumptions; can struggle under severe partial observability and non-stationarity &
COMA: counterfactual baseline for multi-agent credit assignment with centralized critic \\
\hline

Jing et al.~\cite{jing2025fmarlsurvey} &
Federated (Multi-agent + privacy-preserving) &
2025 &
Federated multi-agent reinforcement learning &
State-of-the-art survey anchor; covers privacy, scalability, heterogeneity, and communication costs &
Survey paper; robotics-specific benchmarks remain limited &
FMARL survey of methods, applications, and open challenges for distributed learning \\
\hline

\end{tabular}
\end{table*}

\subsection{Centralized Intelligence: AI in the Cloud}

In centralized architectures, inference and decision-making processes are executed on remote servers, while IoT and robotic nodes act primarily as data collectors and action executors~\cite{pan2020edge, zhang2020empowering_aiot}. Cloud-based intelligence offers access to high-performance computing, global situational awareness, and scalable training/deployment pipelines~\cite{hu2012cloud_robotics}. It also naturally supports platform-centric Internet of Robotic Things (IoRT) integration and orchestration across heterogeneous devices and services~\cite{vermesan2020iort_platforms}.

Applications include fleet-level optimization, large-scale monitoring, and analytics-intensive decision support~\cite{zhou2019edge}. However, centralized intelligence suffers from network dependence and latency variability, which limit suitability for hard real-time loops and safety-critical behaviors. Consequently, centralized AI is commonly complemented with hybrid designs that offload only selected components and place time-sensitive functions closer to the physical system.

\subsection{Decentralized Intelligence: On-Device AI}

Decentralized intelligence embeds AI models directly into edge or robotic devices using specialized hardware (e.g., NPUs/GPUs) and optimized lightweight models~\cite{xu2020edge, zhu2020toward_intelligent_edge}. This allows devices to process sensor data locally and execute decisions without requiring continuous connectivity, which is essential under intermittent networking or safety-critical response constraints.

Recent progress has improved feasibility of local reasoning, including memory-efficient long-context inference for language-enabled modules on constrained platforms~\cite{chen2025edgeinfinite}. In addition, modern compression pipelines (quantization, pruning, distillation, and hybrid approaches) systematize how language and other deep models can be made deployable within tight memory/latency budgets~\cite{zhu2024llm_compression_survey}. Nonetheless, the key trade-off remains limited local compute and energy budgets, which constrain model complexity and adaptation capabilities, especially for multi-modal, multi-robot workloads.

\subsection{Federated Intelligence: Distributed Learning Across Agents}

Federated learning enables multiple robots or edge devices to collaboratively train models without sharing raw data~\cite{lim2020federated}. Each agent trains a local model and periodically shares parameters or gradients with an aggregator, which produces a global model update~\cite{shi2020communication}. This is attractive when privacy, bandwidth, and organizational boundaries prevent centralized data collection.

In robotics, federated learning can support fleet-level improvement while keeping sensitive operational data local. Federated reinforcement learning and federated multi-agent reinforcement learning (FMARL) extend this idea to settings where shared experience can improve coordination strategies among agents~\cite{jing2025fmarlsurvey}. Key challenges include communication overhead, convergence stability under heterogeneous devices, and non-iid data across agents, where performance can degrade if system heterogeneity and data imbalance are not explicitly addressed~\cite{jing2025fmarlsurvey, liu2020survey}.

\subsection{Multi-Agent Learning and Coordination}

In multi-robot systems, agents must coordinate behaviors based on shared goals and environmental feedback~\cite{zhang2019multi}. Decentralized partially observable Markov decision processes (Dec-POMDPs), consensus-based control, and game-theoretic strategies are frequently used~\cite{sutton2018reinforcement, sabattini2013distributed_global_connectivity}. Multi-agent deep reinforcement learning (MADRL) has enabled coordinated behavior in complex environments such as drone swarms, warehouse fleets, and heterogeneous teams.

A widely adopted conceptual anchor for cooperative MARL is centralized training with decentralized execution (CTDE), which improves learning stability while enabling decentralized policies at runtime~\cite{foerster2018coma}. However, CTDE methods can struggle under severe partial observability, non-stationarity, and communication constraints, motivating hybrid learning and communication-aware coordination strategies.

\subsection{LLM-Assisted Decision-Making and SLM-Based Autonomy}

Recent advances in language understanding have enabled robots to interpret and execute high-level instructions using LLMs. Early work by Tellex et al.~\cite{tellex2011understanding} demonstrated grounding of natural language commands into robotic actions, and subsequent work extended language-conditioned grounding and planning for complex manipulation and goal-directed behavior~\cite{misra2016tell, paxton2019prospection}. Recent survey work further systematizes how LLMs are integrated into robotic systems and how they interact with perception, planning, and control modules~\cite{kim2024survey}. Recent studies have also explored LLM-based robotic agents that emphasize goal-directed reasoning and autonomous decision support, while still relying on conventional perception, control, and learning pipelines for real-world execution~\cite{Huang2025AgenticLLMRobotics}.

While cloud-hosted LLMs can support semantic understanding and high-level task inference, their size and latency pose challenges for real-time decision-making on resource-constrained platforms. This motivates increasing interest in SLMs and device-deployable language modules. Practical progress includes memory-efficient long-context inference on edge devices~\cite{chen2025edgeinfinite} and systematically developed compression pipelines for deployable language models~\cite{zhu2024llm_compression_survey}. Integrating such models into decentralized architectures supports responsive autonomy without constant cloud connectivity, particularly when combined with robust communication substrates and orchestration strategies for hybrid deployments (e.g., ROS~2 middleware selection and latency-aware edge--cloud scheduling)~\cite{zhang2024ros2_middleware, sun2025latencyaware}.

Integrating SLMs into decentralized architectures offers a promising direction for scalable, interpretable, and responsive autonomous systems capable of reasoning and coordination in dynamic environments.

\subsection{Ethical and Social Considerations}

The large-scale integration of AI, IoT, and robotic systems raises several
ethical and societal challenges that must be addressed to ensure safe and
responsible deployment. A primary concern involves privacy and data
protection, as pervasive IoT sensing and robot-mounted cameras can expose
sensitive information about individuals, environments, and operational
contexts. Recent work on privacy-preserving perception, such as the use of
region-of-interest masking for anonymizing learned image compression
\cite{10711721}, highlights the need for dedicated mechanisms that filter or
transform sensor data before storage or transmission, particularly in
surveillance and disaster-response applications. The distributed nature of
IoT–robotic networks further increases the attack surface for cyber–physical
intrusions, requiring strong authentication, encryption, and secure
data-handling mechanisms across edge, fog, and cloud layers~\cite{roman2013features}.

Security concerns are closely linked to the autonomy of robotic systems
operating under partial observability or unreliable network conditions.
Compromised or spoofed sensor data, model poisoning, or adversarial AI inputs
can lead to unsafe decisions, especially in mission-critical applications such
as disaster response, healthcare assistance, or industrial automation.
Recent work on secure federated learning has shown that communication-layer
vulnerabilities can further expose distributed robotic systems to privacy leakage
and adversarial interference, and that emerging 6G mechanisms, such as
intelligent reflecting surfaces, can enhance secrecy rates and strengthen
defense against such threats~\cite{mao2025blockchain_coldstart_frl_leo, mao2025secrecy, DBLP:journals/comsur/MaoTKK22}. Ensuring robustness against
these risks is essential for maintaining operational safety and trustworthiness.

Human–AI interaction also introduces important ethical questions related to
transparency, accountability, and shared control. Systems that rely on AI
decision-making must provide interpretable and predictable behaviors to human operators to enable effective oversight, especially when robots perform tasks that may affect human well-being. Clear guidelines for responsibility and
fallback behaviors are needed when humans and robots collaborate in dynamic settings~\cite{winfield2019ethical}.

In summary, the deployment of AI–IoT–Robotic systems requires not only
technical advancement but also careful consideration of privacy, security,
and human-centered design principles. These issues remain open research
frontiers that will shape the practical and societal impact of future
intelligent infrastructures.

\section{Challenges and Future Directions}

\begin{table*}[!t]
\centering
\caption{Detailed Survey of Emerging Trends: Paper-Level Contributions (Updated with Recent Anchors)}
\label{tab:emerging_trends_paperwise}
\begin{tabular}{|p{2.6cm}|p{0.7cm}|p{3.0cm}|p{5.0cm}|p{5.0cm}|}
\hline
\textbf{Citation} & \textbf{Year} & \textbf{Trend Category} & \textbf{Key Contribution} & \textbf{Limitations / Open Challenges} \\
\hline

\multicolumn{5}{|c|}{\textbf{TinyML and Model Compression Trends}} \\
\hline

Banbury et al.~\cite{banbury2021mlperftiny} & 2021 &
TinyML benchmarking &
Introduces MLPerf Tiny, an industry-aligned benchmark suite measuring accuracy, latency, and energy for TinyML inference on ultra-low-power devices &
Benchmark tasks are not robotics-specific; closed-loop autonomy metrics (control stability, actuation timing, safety) and multi-sensor pipelines are not directly evaluated. \\
\hline

Abadade et al.~\cite{abadade2023tinymlsurvey} & 2023 &
TinyML landscape and tooling &
Comprehensive survey of TinyML methods (compression, efficient architectures, deployment stacks) and evaluation practices, providing an updated taxonomy and research map &
Limited treatment of embodied robotics constraints (real-time control loops, runtime verification) and distributed multi-robot updates under communication constraints. \\
\hline

Zhu et al.~\cite{zhu2023modelcompressionsurvey} & 2023 &
Modern model compression survey &
Recent survey consolidating pruning, quantization, distillation, and architecture-level efficiency, with emphasis on deploying DNNs on constrained devices &
Robotics-specific closed-loop requirements (stability, safety margins, schedulability) and field-deployment evaluation remain underrepresented. \\
\hline

\multicolumn{5}{|c|}{\textbf{Edge Intelligence and Small Language Models}} \\
\hline

Chen et al.~\cite{chen2025edgeinfinite} & 2025 &
Efficient long-context Transformer inference &
Proposes memory-efficient Transformer inference for longer-context workloads on constrained devices, targeting practical edge execution &
Robotics embodiment effects (sensor noise, contact dynamics) and system-level integration (perception--planning--control under deadlines) are not the primary validation focus. \\
\hline

Chitty-Venkata et al.~\cite{chittyvenkata2023transformerinference} & 2023 &
Transformer inference optimization (survey) &
Survey of techniques for optimizing Transformer inference, covering algorithmic and systems-level methods (memory efficiency, kernel/serving optimizations, quantization, and sparsity) &
Survey-level: does not establish embodied robotics benchmarks; robustness, real-time guarantees, and safety under closed-loop interaction remain open. \\
\hline

Qu et al.~\cite{qu2025jointprunequant} & 2025 &
Joint pruning + quantization (modern compression) &
Introduces an automated framework for jointly optimizing structured pruning and quantization, improving practical compressibility and training/deployment efficiency &
Results are largely DNN-centric; translating joint compression decisions into robotics stacks (real-time constraints, perception drift, safety envelopes) requires further validation. \\
\hline

\multicolumn{5}{|c|}{\textbf{Adaptive and Self-Reconfigurable Learning}} \\
\hline

Kumar et al.~\cite{kumar2021rma} & 2021 &
Real-time online adaptation &
Proposes Rapid Motor Adaptation (RMA) for fast online adaptation in legged robots, demonstrating real-time robustness to unseen terrain and perturbations &
Primarily locomotion-focused; generalizing to heterogeneous IoT sensing and broader task classes (manipulation, inspection) and formal safety guarantees remains open. \\
\hline

Yousefi et al.~\cite{yousefi2025sharedautonomy} & 2025 &
Human-in-the-loop policy alignment &
Presents a shared autonomy framework for human-in-the-loop policy fine-tuning and alignment for robotic tasks, supporting practical adaptation in interactive settings &
Human-in-loop availability, user variability, and integration with heterogeneous IoT sensing and multi-robot coordination under communication constraints remain challenging. \\
\hline

Ren et al.~\cite{ren2023hippomat} & 2023 &
Dynamic multi-robot reconfiguration &
Presents hierarchical decentralized matching for scalable multi-robot task allocation under changing demands, supporting reconfiguration under dynamic workloads &
Assumptions and benchmarks may differ from disaster/field deployments; resilience under severe bandwidth limits, intermittent connectivity, and sensing uncertainty remains critical. \\
\hline

Chakraa et al.~\cite{chakraa2023mrtaoptreview} & 2023 &
MRTA (updated survey anchor) &
Provides a recent review of optimization-based MRTA strategies, including taxonomies, trends, and open research directions relevant to dynamic environments &
Survey emphasizes optimization methods; explicit coupling to modern edge/IoT constraints (distributed compute, on-device learning, semantic reasoning) remains an open systems challenge. \\
\hline

\multicolumn{5}{|c|}{\textbf{Cloud, Human-in-the-Loop, and Hybrid Collaboration}} \\
\hline

Groshev et al.~\cite{groshev2023edgerobotics} & 2023 &
Edge robotics evaluation (cloud--edge continuum) &
Experimental evaluation of edge/fog robotics readiness, highlighting practical benefits and bottlenecks of current network/virtualization technologies for robotics workloads &
Findings are application- and stack-dependent; general benchmarks for heterogeneous robots, safety constraints, and harsh-field conditions remain incomplete. \\
\hline

Zhang et al.~\cite{zhang2024ros2_middleware} & 2024 &
ROS\,2 edge-to-cloud communication &
Compares middlewares for distributed ROS\,2 systems across edge-to-edge and edge-to-cloud settings, strengthening evidence-based design choices for hybrid robotic deployments &
Performance depends on hardware/network configurations; broader validation for multi-robot scaling, security, and real-time QoS in field conditions remains needed. \\
\hline

Sun et al.~\cite{sun2025latencyaware} & 2025 &
Latency-aware IoT--edge--cloud scheduling &
Proposes latency-aware scheduling for data-oriented service requests in collaborative IoT-edge-cloud networks, improving response time and load balancing under task dependencies &
Assumes modelable latency/traffic; robustness under highly dynamic wireless conditions, failures, and real robotic sensing/actuation loops requires further study. \\
\hline

Sahoo and Mishra~\cite{sahoo2025edgescheduling} & 2025 &
Edge-cloud task scheduling (survey anchor) &
Survey of task scheduling in edge-cloud systems, organizing scenarios, methods, and trade-offs relevant to collaborative scheduling in hybrid infrastructures &
Survey-level: does not resolve robotics-specific constraints such as safety certification, deadline-driven control, and multi-robot coordination with semantic workloads. \\
\hline

\end{tabular}
\end{table*}

Despite notable progress at the intersection of AI, IoT, and Robotics, several barriers continue to impede seamless integration, real-world deployment, and scalable generalization. This section consolidates key research challenges and identifies priority directions to shape the next decade of development. Table~\ref{tab:emerging_trends_paperwise} presents a paper-level breakdown of representative contributions aligned with each trend discussed, while Table~\ref{tab:intelligence_models} complements it by contrasting how intelligence is placed and coordinated across cloud-centric, hybrid, on-device, and federated paradigms.

\subsection{Security, Privacy, and Trust}

The convergence of robotics, pervasive IoT sensing, and cloud- and edge-hosted AI expands the system attack surface and introduces new risks across communication, computation placement, and learning pipelines. Beyond classical threats (adversarial inputs, spoofing, leakage)~\cite{zafir2024security, zhou2019edge}, emerging deployments add additional exposure through hybrid edge--cloud orchestration and distributed middleware layers used in practice~\cite{zhang2024ros2_middleware}. Future efforts must design trustworthy AI--IoT--robotic systems that ensure verifiable security under continuous operation and protect privacy, particularly for fleet-scale learning and collaboration where updates and coordination occur across heterogeneous devices~\cite{price2017ai_healthcare_legal, jing2025fmarlsurvey}. While encryption and blockchain-based authentication are emerging~\cite{fernandez_carames2018blockchain_iot_review}, real-time threat detection, secure-by-design orchestration, and resilience under intermittent connectivity remain open problems.

\subsection{Interoperability and Standardization}

Fragmented protocols between robotic middleware and IoT communication stacks remain a core integration bottleneck~\cite{liu2020survey, al2015internet}. Disparities in data formats, transport layers, and semantic representations hinder interoperability and cross-platform deployment~\cite{datta2018interoperable_smarthome}. Recent evidence-based studies comparing communication substrates for distributed ROS~2 systems highlight how middleware choices affect end-to-edge and edge-to-cloud performance, and therefore must be treated as a first-class design variable in standardization efforts~\cite{zhang2024ros2_middleware}. A clear direction for future work involves standard APIs, message ontologies, and real-time bridges that unify AI inference, IoT telemetry, and robotic control pipelines, together with reference configurations that account for hybrid service placement and scheduling across IoT--edge--cloud tiers~\cite{carreira2024ros2_gateway, sun2025latencyaware, lu2017industry}.

\subsection{Scalability and Latency in Multi-Agent Intelligence}

Intelligent robotic systems increasingly operate as distributed fleets, motivating scalable multi-agent learning and coordination under partial observability and communication constraints~\cite{foerster2018coma, gronauer2022marl_survey}. Scaling such systems requires synchronizing distributed knowledge, minimizing inference latency, and controlling real-time bandwidth usage~\cite{zhou2019edge, xu2020edge}. Recent progress toward on-device language-enabled autonomy improves feasibility for decentralized reasoning under memory budgets~\cite{chen2025edgeinfinite}, while systematic compression pipelines provide practical guidance for deployable language models on constrained platforms~\cite{zhu2024llm_compression_survey}. At the systems level, hybrid deployments demand latency-aware scheduling of dependent services across edge and cloud tiers~\cite{sun2025latencyaware} and robust middleware choices for distributed robot networks~\cite{zhang2024ros2_middleware}. Further research should evaluate integrated approaches that combine CTDE/MADRL foundations with federated multi-agent learning for fleet-scale adaptation under heterogeneity and non-iid data~\cite{jing2025fmarlsurvey}.

\subsection{Learning Beyond Static Datasets}

Robotic AI pipelines are predominantly trained on pre-collected datasets, limiting adaptability in open-ended environments~\cite{zhang2020empowering_aiot, ando2021iort_aal_patterns}. Few benchmarks provide synchronized streams of multimodal IoT signals, robot interaction traces, and semantic supervision suitable for evaluating closed-loop generalization. There is a need for open benchmarks and simulation-to-real testbeds that jointly evaluate robustness, safety, and responsiveness under realistic latency and compute budgets. Future work should prioritize interaction-centric learning paradigms, online, continual, and reinforcement learning in-the-loop, and explicitly study how updates propagate across cloud, edge, and federated settings for heterogeneous fleets~\cite{jing2025fmarlsurvey, sun2025latencyaware}.

\subsection{Architectural Integration and Reference Designs}

Most AI--IoT--robotic systems remain ad hoc, lacking shared reference designs that specify how sensing, reasoning, actuation, and adaptation interact across edge/fog/cloud tiers. The field requires modular architectures that make explicit (i) platform-level IoRT integration and orchestration concepts~\cite{vermesan2020iort_platforms}, (ii) communication substrates and middleware configurations for distributed robot systems~\cite{zhang2024ros2_middleware}, and (iii) service placement and scheduling policies that preserve responsiveness under dynamic workloads~\cite{sun2025latencyaware}. Reference designs should formalize closed-loop feedback between IoT perception, AI cognition, and robotic actuation, while providing validated operating envelopes for latency, bandwidth, and failure modes.

\subsection{Roadmap and Long-Term Research Priorities}

Based on the gaps outlined above, we recommend the following roadmap over the next 5--10 years:

\begin{itemize}
    \item \textbf{2025--2027:} Establish reference stacks and reproducible testbeds for AI--IoT--robotics integration, including standardized middleware configurations for distributed ROS~2 deployments, and latency-aware orchestration across IoT--edge--cloud tiers~\cite{zhang2024ros2_middleware, sun2025latencyaware}. Expand open benchmarks combining multimodal IoT sensing, robot interaction traces, and evaluation of closed-loop timing constraints.
    \item \textbf{2027--2030:} Advance trustworthy and privacy-preserving fleet learning, including federated multi-agent reinforcement learning under heterogeneity and non-iid data, and develop security-by-design orchestration and monitoring methods for distributed robotic infrastructures~\cite{jing2025fmarlsurvey}.
    \item \textbf{2030 onward:} Transition toward lifelong adaptive agents capable of self-reconfiguration and mission-level decision making, supported by deployable on-device reasoning modules (including compressed language-enabled components) and validated hybrid architectures that operate robustly under intermittent connectivity and uncertain environments~\cite{zhu2024llm_compression_survey, chen2025edgeinfinite}.
\end{itemize}

Although a detailed discussion of 6G-enabled pervasive intelligence and extraterrestrial robotic missions lies outside the scope of this survey, the unified perception–cognition–action framework and distributed intelligence models presented here provide conceptual foundations that can inform future research in such emerging domains.

\section{Conclusion}

The convergence of Artificial Intelligence, Internet of Things, and Robotics is reshaping the foundations of autonomy across domains. This survey has examined their integration from multiple perspectives, taxonomical, architectural, application-driven, and intelligence-distribution-focused perspectives. It has also identified the systemic challenges impeding their unification and proposed a consolidated research agenda.

As these domains continue to evolve, the demand for intelligent, context-aware, and cooperative systems will increase. Realizing this vision requires more than engineering bridges between silos; it necessitates a fundamental rethinking of how perception, reasoning, and actuation can be distributed, coordinated, and continuously improved across physical and digital substrates.

Ultimately, the fusion of AI, IoT, and Robotics is not merely a technical integration challenge; it is a scientific opportunity to redefine autonomy itself. We hope this survey serves as both a roadmap and a call to action for the research community to embrace this intersection as a fertile ground for innovation and impact.

\subsection{Vision of the Future: Toward Connected Robotics and Physical AI}

While this survey has synthesized current approaches to AI–IoT–Robotics integration, we believe the next decade will mark a decisive shift toward Connected Robotics and Physical AI. Unlike traditional multi-robot systems, which emphasize localized cooperation for task execution, connected robotics views robots as persistent nodes in a global cyber-physical fabric, continuously linked with IoT infrastructure, humans, and cloud intelligence. This paradigm enables autonomy that is not only collaborative but also infrastructural, embedding robots into the daily operation of smart cities, healthcare systems, logistics, and environmental management.

At the same time, the notion of Physical AI will emerge as a central research theme. If today’s AI excels in symbolic reasoning and language, the next generation must demonstrate intelligence through embodiment: robots that learn, adapt, and reason in real-world environments under uncertainty. Achieving this vision requires advances in interoperability, self-reconfigurable architectures, distributed cognition, and ethical alignment.

We foresee that within ten years, the unification of connected robotics and physical AI will transform autonomous systems from specialized tools into a shared infrastructure, analogous to the role the Internet plays today, fundamentally redefining how society interacts with intelligent machines.

\bibliographystyle{IEEEtran}
\bibliography{IEEEabrv,root}



\begin{IEEEbiography}
        [{\includegraphics[width=1in,height=1.25in,clip,keepaspectratio]{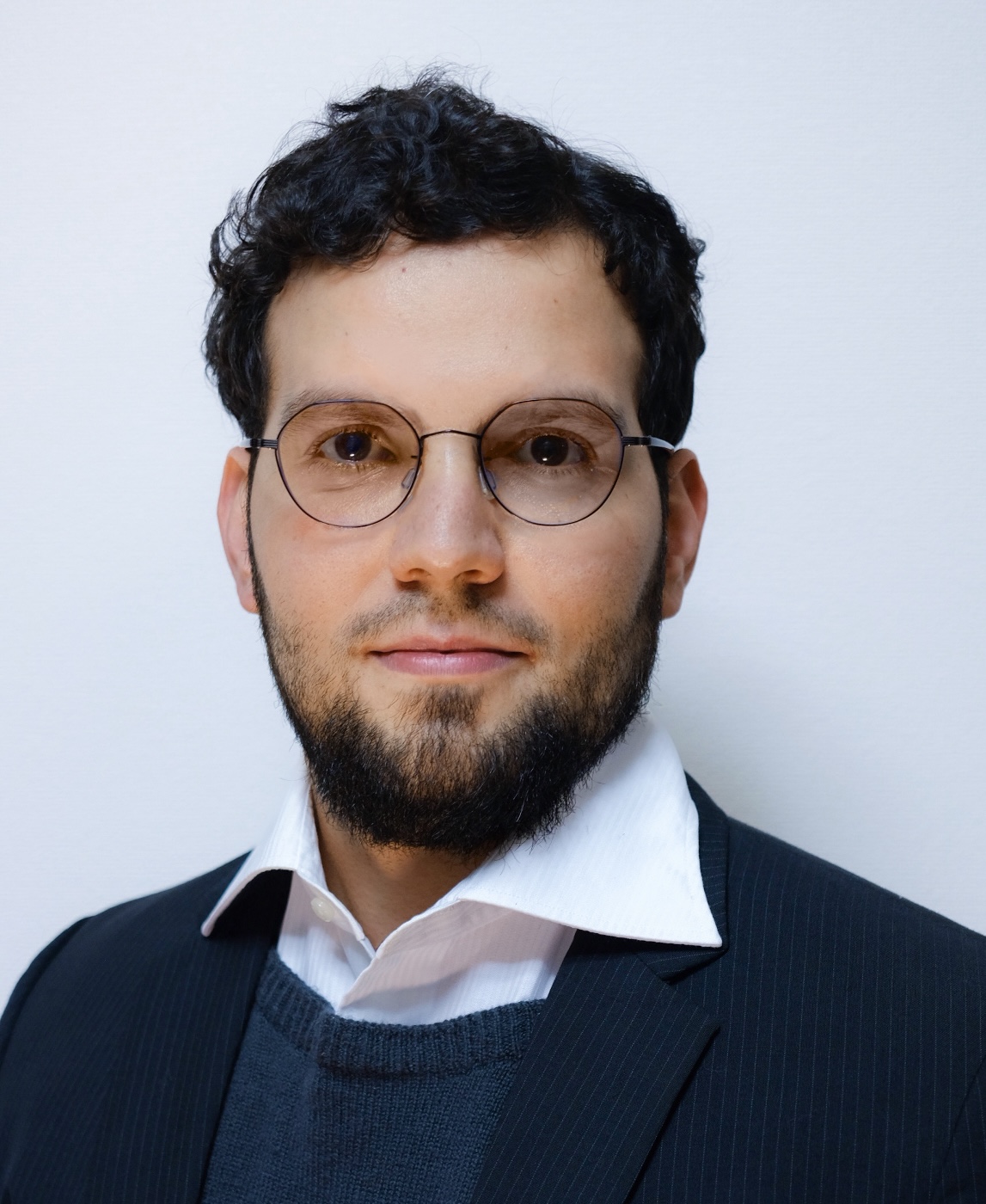}}]{Ranulfo Bezerra}
        received his Ph.D. degree at Tohoku University, Japan in 2021. He
        received his M.Sc. and B.Sc. in computer science from the Federal
        University of Piaui, Brazil in 2018, and 2016, respectively. He is
        currently an assistant professor at the Tough
        Cyberphysical AI Research Center (TCPAI) at Tohoku University. His research
        interests are intelligent robotic systems, robotic perception and
        autonomous robotic systems and their related applications. A member of RSJ
        and IEEE.
\end{IEEEbiography}
\vspace{11pt}
\begin{IEEEbiography}
        [{\includegraphics[width=1in,height=1.25in,clip,keepaspectratio]{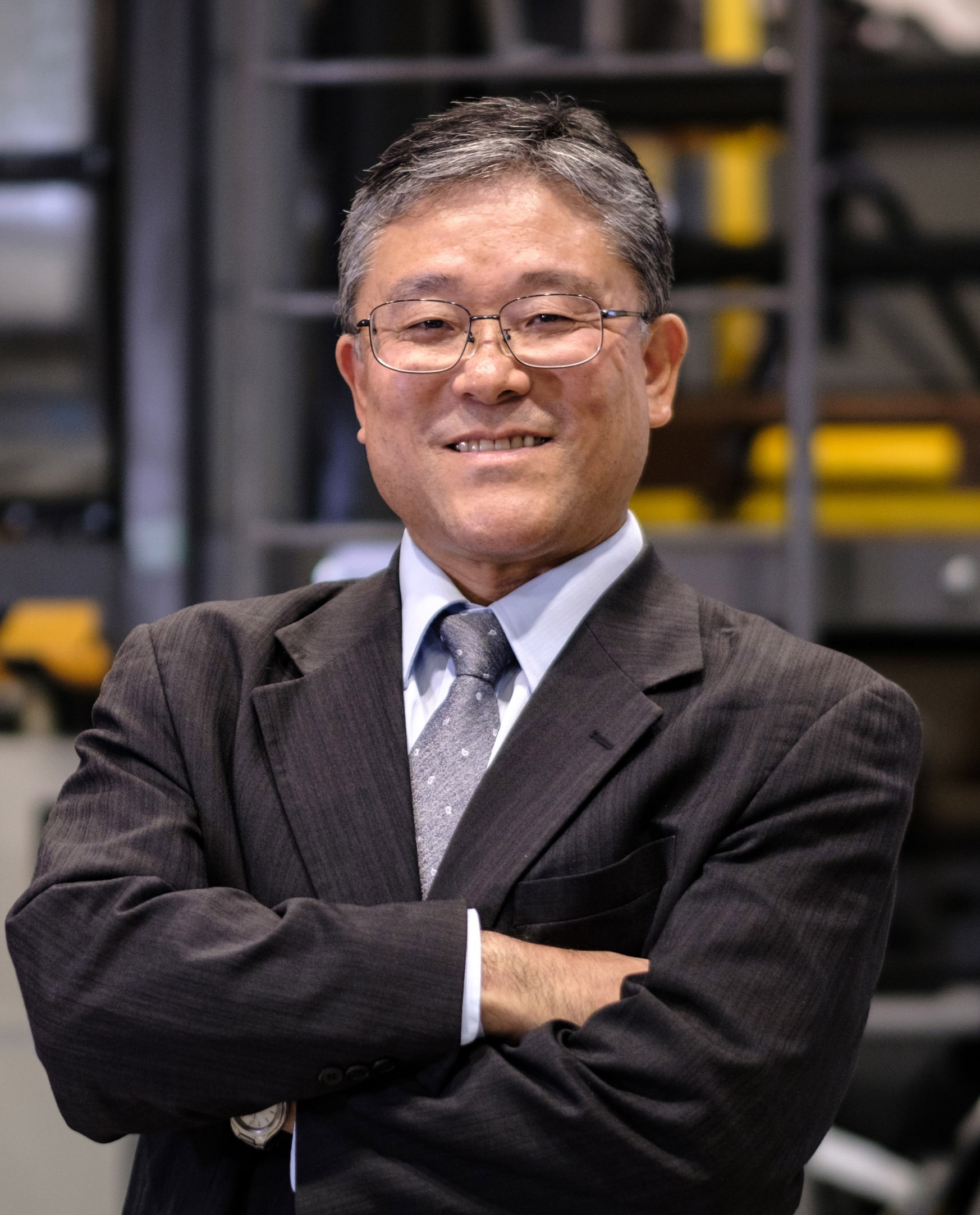}}]{Satoshi Tadokoro}
        (M'89 - SM'06 - F'09) graduated from the University of Tokyo in 1984. He
        was an associate professor in Kobe University in 1993-2005, and has been
        a Professor of Tohoku University since 2005. He was a Vice/Deputy Dean of
        Graduate School of Information Sciences in 2012-14, and is the Director
        of Tough Cyberphysical AI Research Center since 2019 in Tohoku
        University. He has been the President of International Rescue System Institute
        since 2002, and was the President of IEEE Robotics and Automation
        Society in 2016-17. He received awards including IEEE RAS George Saridis
        Leadership Award in Robotics and Automation, and Commendation for S$\&$T
        by the Minister of MEXT Prize for S$\&$T. He served as the Program
        Manager of MEXT DDT Project on rescue robotics in 2002-07, and was the Project
        Manager of Japan Cabinet Office ImPACT Tough Robotics Challenge Project on
        disaster robotics in 2014-19 having 62 international PIs and 300 researchers
        that created Cyber Rescue Canine, Dragon Firefighter, etc. His research
        team in Tohoku University has developed various rescue robots, two of which
        called Quince and Active Scope Camera are widely recognized for their contribution
        to disaster response including missions in the Fukushima-Daiichi NPP nuclear
        reactor buildings. IEEE Fellow, RSJ Fellow, JSME Fellow, and SICE Fellow.
    \end{IEEEbiography}
\vspace{11pt}
\begin{IEEEbiography}
        [{\includegraphics[width=1in,height=1.25in,clip,keepaspectratio]{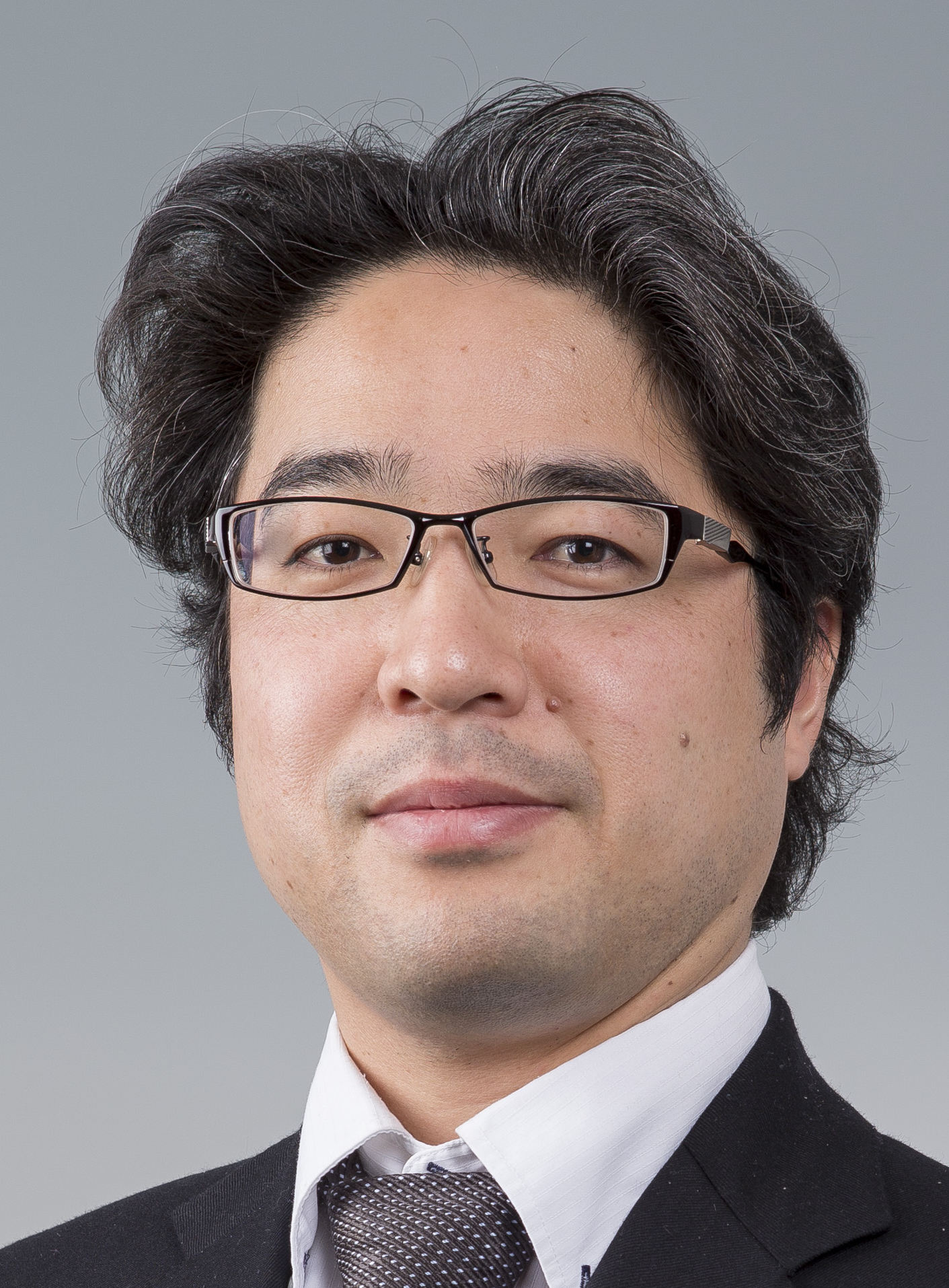}}]{Kazunori Ohno}
        received BS, MS, and Ph.D. from Tsukuba University in 1999, 2001, and 2004.
        He was a COE researcher at Kobe University in 2004, became an assistant
        professor, a lecturer, and an associate professor at Tohoku University
        in 2005, 2008, and 2012, and has been a professor of NICHe Tohoku University
        since 2021. He was also a PRESTO researcher (2008-2012) and has been a visiting
        researcher of the RIKEN Center of AIP (2017-2022). His research fields
        are field robotics, robot intelligence, and cyber-enhanced canines. He established
        RC on Data Engineering Robotics of RSJ in 2012. He received Kisoi awards
        in 2008 and 2012, RSJ research awards in 2005 and 2019. A member of RSJ,
        JSME, VRSJ, JSAE, IEEE.
\end{IEEEbiography}

\vfill

\end{document}